\documentclass[11pt]{article}
\usepackage{rotating}
\usepackage{acl2015}
\usepackage{times}
\usepackage{url}
\usepackage{latexsym}
\usepackage{tabularx}
\usepackage{makecell}
\usepackage{hyperref}
\usepackage{multirow}
\usepackage{multicol}
\usepackage{booktabs}
\usepackage{appendix}
\usepackage{CJKutf8}
\usepackage[section]{placeins}
\usepackage[hang,flushmargin]{footmisc} 
\usepackage[T1]{fontenc}
\usepackage{lipsum}
\newcommand\blfootnote[1]{%
  \begingroup
  \renewcommand\thefootnote{}\footnote{#1}%
  \addtocounter{footnote}{-1}%
  \endgroup
}
\usepackage{fancyhdr}
\usepackage{caption}

\title{ChatHome: Development and Evaluation of a Domain-Specific Language Model for Home Renovation}

\author{Cheng Wen, Xianghui Sun, Shuaijiang Zhao,\\
\textbf{Xiaoquan Fang, Liangyu Chen, Wei Zou}\\
Beike Inc., Beijing, China  \\
\texttt{\{wencheng008,sunxianghui002,zhaoshuaijiang001,} \\
\texttt{fangxiaoquan001,chenliangyu003,zouwei026\}@ke.com}}

\begin{document}
\maketitle
\begin{abstract}
This paper presents the development and evaluation of ChatHome, a domain-specific language model (DSLM) designed for the intricate field of home renovation.
Considering the proven competencies of large language models (LLMs) like GPT-4 and the escalating fascination with home renovation, this study endeavors to reconcile these aspects by generating a dedicated model that can yield high-fidelity, precise outputs relevant to the home renovation arena. ChatHome's novelty rests on its methodology, fusing domain-adaptive pretraining and instruction-tuning over an extensive dataset. This dataset includes professional articles, standard documents, and web content pertinent to home renovation. This dual-pronged strategy is designed to ensure that our model can assimilate comprehensive domain knowledge and effectively address user inquiries. Via thorough experimentation on diverse datasets, both universal and domain-specific, including the freshly introduced "EvalHome" domain dataset, we substantiate that ChatHome not only amplifies domain-specific functionalities but also preserves its versatility.
\end{abstract}

\section{Introduction}

In the vibrant arena of artificial intelligence, the development of large-scale language models such as GPT4\cite{gpt4} and ChatGPT\cite{gpt3.5} has triggered profound shifts in natural language processing tasks, exhibiting an impressive degree of competence in myriad tasks. 
Chinese open-source large language models also grow rapidly, such as ChatGLM\cite{glm130b}, Baichuan\cite{Baichuan13Bgithub}, BELLE\cite{BELLE} .
While some works have made notable strides in several domains, including healthcare\cite{wang2023huatuo}, finance\cite{wu2023bloomberggpt}, and law\cite{huang2023lawyer}\cite{cui2023chatlaw}, the specific domain of home renovation remains relatively unexplored. 

Home renovation is a multifaceted field that demands a comprehensive grasp of both aesthetics and functionality. It extends beyond mere selection of furniture or color palette identification. Rather, it necessitates a profound understanding of architectural nuances, spatial design principles, human-centric design considerations, and prevailing trends, among other elements.


However, mainstream models like ChatGPT, despite their general capabilities in various tasks, often fail to generate domain-specific content with high fidelity and accuracy, as observed in the previous examples of legal\cite{huang2023lawyer,cui2023chatlaw} and medical \cite{wang2023clinicalgpt,wang2023huatuo}domains. Therefore, to overcome these shortcomings and to cater to the unique demands of the home renovation sector, there is an urgent need for a specialized language model tailored to this domain.

This study proposes ChatHome, a language model specifically designed for home renovation. Our approach involves two steps: first, post-pretraining a generic model using a wide-ranging home renovation dataset, encompassing professional articles, standard documents, and web content. Second, implementing an instruction-tuning strategy with a dataset of question-answer pairs, generated using home renovation-based prompts. 

This research aims to show that post-pretraining and fine-tuning large language models improves their performance in specific domains. While enhancing the capabilities of specific domains, we also pay attention to the changes in the general capabilities of the model and conduct detailed evaluations, which will be described in detail in subsequent sections.

In summary, there are two main contributions of this paper:

\begin{itemize}
\item 
We established ChatHome, a fine-tuned LLM focused on the home renovation domain.
\item
We introduced a domain dataset and conducted comprehensive experiments on both universal and domain datasets to verify the effectiveness of our model.
\end{itemize}

\section{Related work}
The training of a LLM usually includes two stages: pre-training and instruction fine-tuning. Through pre-training on a large-scale corpus, the LLM can obtain basic language understanding and generation capabilities. The instruction fine-tuning stage is to enable the model to have the ability to understand human instructions, and can also improve the generalization ability of the model on unseen tasks\cite{ouyang2022training}\cite{zhao2023survey}.
However, domain-specific tasks often involve complex concepts, technical terminology, and complex relationships between entities\cite{ling2023domain}. Without targeted guidance, large language models can severely hallucinate. This occurs because LLMs aim to predict the most likely sequence of words given an input, rather than provide a definitive answer based on structured knowledge. 

Recently, a lot of work related to large language adaptation has emerged in the fields of medical\cite{wang2023huatuo}, financial\cite{wu2023bloomberggpt}\cite{yang2023fingpt} and legal field\cite{cui2023chatlaw}\cite{huang2023lawyer}.
Using the retrieval-based plug-in knowledge base, LLM can be used in professional fields without updating parameters\cite{ram2023incontext}, or you can choose to inject domain knowledge into the model by updating parameters. This report mainly focuses on the latter. 

LLM field specialized training methods can be roughly divided into the following categories according to different training stages: one method is to pre-train from scratch directly based on domain data, such as \cite{wu2023bloomberggpt}, which usually relies on a large amount of domain data, and the training cost is high; one is to perform fine-tuning directly based on domain instruction data, such as \cite{cui2023chatlaw}\cite{wang2023huatuo}; and the other is to perform domain pre-training on the foundation LLM based on domain data, and then perform instruction fine-tuning\cite{huang2023lawyer}.

\section{Data Collection}

\subsection{Pre-training Corpus}
\label{Pre-trainingCorpus}

Prior research\cite{lee2020biobert} has shown that language models can benefit from the knowledge acquired through domain-specific corpora. We gather a domain-specific corpus to enhance the models with knowledge about home decoration. Additionally, we compile a general corpus to provide the models with a balance of general knowledge.

\textbf{National Standards.}
We collect several National Standards for decoration and construction, of which worth mentioned are
Code for design of residential buildings(GB 50096-2011), Code for construction of decoration of housings(GB 50327-2001).

\textbf{Domain Books.} 
We collect books in the field of real estate, home renovation, decoration and construction, which published in the past decade.

\textbf{Domain Websites.} 
We crawl domain websites, about 30,000 articles on the category of home renovation advice, home devices purchasing skills and so on.

\textbf{General Corpus.} 
 To construct the general corpus, we sample articles from WuDaoCorpora\cite{yuan2021wudaocorpora}, the Simplified Chinese edition of Wikipedia.

\begin{figure*}[htbp]
\centering
\setlength{\abovecaptionskip}{0.1cm} 
\includegraphics[width=0.9\textwidth, trim={1cm 7cm 1cm 7cm} ]{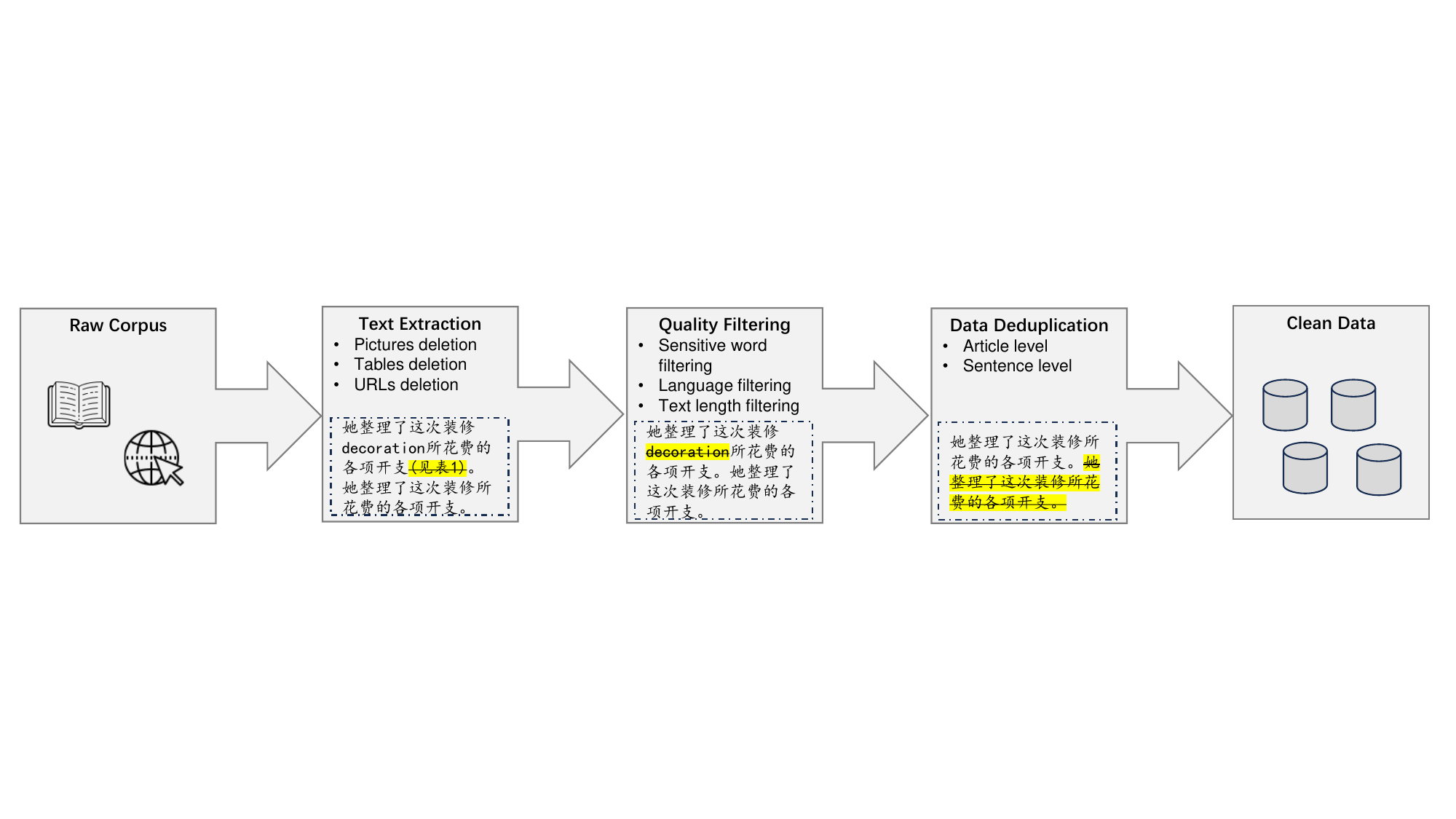}
\caption{Workflow of data preprocessing.}
\label{data_process}
\end{figure*}

\textbf{Data Preprocessing.}
The above data is processed through a unified pipeline, consisting of text extraction, quality filtering, and data deduplication. 
During text extraction, we discard irrelevant information like pictures, tables, and URLs, only storing the relevant text. 
During quality filtering, we ensure that each data is usable through methods like sensitive word filtering, language filtering, and effective text length filtering. 
Furthermore, we minimize the influence of duplicate data on model training by deduplicating at both the article and sentence levels.
Ultimately, we acquire approximately 26.6M tokens from the domain corpus and 276.6M tokens from the general corpus.
The workflow of data preprocessing is shown in Figure\ref{data_process}.

\begin{figure*}[hbt]
\vspace{0.1cm}
\centering
\setlength{\abovecaptionskip}{0.1cm} 
\includegraphics[scale=1]{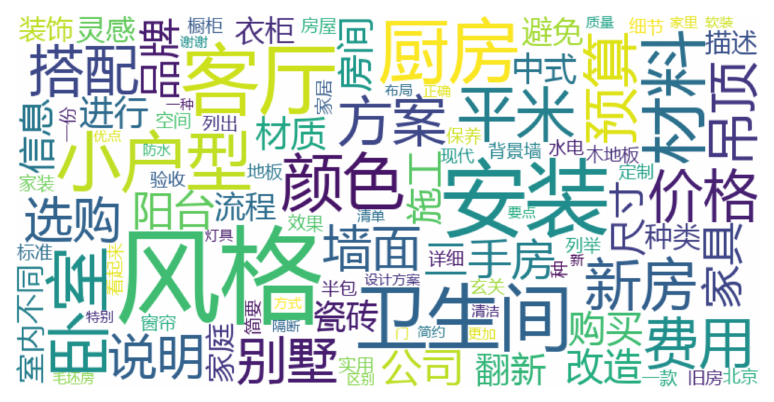}
\caption{Word cloud of instruction data}
\label{Home Decoration keywords}
\end{figure*}

\subsection{SFT Corpus}
In order to alleviate the problem of domain bias and improve the performance of the model in specific fields, 
we construct approximately 25k instruction data from high-quality home decoration books and home decoration website articles to assist the model adapting to specific domain knowledge.
The details about these prompts are introduced in Table\ref{prompts} of Appendix.

\textbf{One-turn Dialogues.}
To acquire more home decor-related questions,
initially, we employ GPT-4 to emulate the dual roles of an interior designer and a client,	
generating several question-answer pairs grounded on the given knowledge.
Subsequently, for obtaining more detailed responses, we directly present the aforementioned questions to GPT-4.	
This two-step approach allows us to gain more comprehensive and precise data.	

\textbf{Multi-turn Dialogues.}
Analogous to one-turn dialogues, GPT-4 emulates the roles of an interior designer and client, facilitating the generation of multi-turn dialogues in the realm of home decoration.	
Furthermore, to mitigate hallucinations, we equip GPT-4 with a pertinent article, thereby orienting its dialogue content around this provided knowledge.	
Moreover, we instruct GPT-4 to maintain focus and handle the dialogue organically.	

Based on the instruction data of one-turn and multi-trun dialogues, we generate a word cloud shown in Figure \ref{Home Decoration keywords}. 

\section{Experiments}

\subsection{Baseline Models}
The baseline model chosen for our study is Baichuan-13B\cite{Baichuan13Bgithub}, which was developed and released by Baichuan Intelligent Technology. There are a total of two models in this baseline.

\textbf{Baichuan-13B-Base} \quad Baichuan-13B-Base is a pre-trained model, boasting a parameter size of 13 billion and a training corpus comprising 1.4 trillion tokens.

\textbf{Baichuan-13B-Chat} \quad Baichuan-13B-Chat, built on the foundational architecture of Baichuan-13B-Base, has been fine-tuned using specialized instructions. Consequently, it demonstrates improved dialogue generation and instruction comprehension capabilities.

\subsection{Experiments Setups}

We apply the aforementioned two baseline models for fine-tuning our home decoration domain dataset. To explore the advantages of domain-adaptive pretraining(DAPT) \cite{DAPTpaper} in domain adaptation, we will conduct identical instruction tuning experiments on models refined using DAPT.

Domain adaptation inevitably confronts the issue of catastrophic forgetting, characterized by the loss of previously acquired knowledge during adaptation to new domains.
One straightforward approach to mitigate this issue is the rehearsal-based strategy, which involves revisiting and relearning previously acquired knowledge. 
Considering that large language models are pre-trained on extensive general-purpose data, achieving a balance between general and domain-specific data during domain adaptation is imperative.
For each experiment, we executed five sets of data ratio tests to determine the most effective data ratio scheme\textsuperscript{1}\blfootnote{
    \textsuperscript{1}The ratio between domain-specific data and general data is 1:0, 1:1, 1:2, 1:5, and 1:10, respectively. The ratio of 1:0 indicates that we exclusively utilized domain-specific data without incorporating any general data
}.

The parameter configuration of the DAPT and SFT stages is shown in Table\ref{hyper-parameters-ft}, the sole distinction in training hyperparameters between DADT and SFT stages lies in the maximum length, with DAPT set to 1024 and SFT to 1536.

\begin{table}[t!]
\caption{Hyper-parameter settings}
\begin{center}
\begin{tabular}{l|c} 
\hline 
\textbf{Hyper parameter} & \textbf{Value} \\
\hline   
Precision  & fp16 \\
\hline
Epochs  & 4 \\
\hline
Batch size  & 64 \\
\hline
Learning rate  & 1e-4 \\
\hline
Warmup ratio  & 0.1 \\
\hline
LR scheduler type  & cosine \\
\hline
\end{tabular}
\end{center}
\label{hyper-parameters-ft}
\end{table}


\subsection{Metrics}
Evaluation is of paramount prominence to the success of LLMs. For ChatHome, we not only hope to inject domain-related knowledge into the model, but also pay attention to the general capabilities of the model after domainization, so our evaluation includes two parts: general capability evaluation and domain capability evaluation.

\textbf{General Evaluation.}  
To evaluate the models on general ability, we adopt C-Eval\cite{huang2023c} and CMMLU\cite{li2023cmmlu}, 
which both are benchmarks evaluating the advanced knowledge and abilities of foundation models in a Chinese context\textsuperscript{2}\blfootnote{
\textsuperscript{2}
We assessed the models in both zero- and few-shot settings, reporting the setup yielding the highest overall average accuracy. Note that, the results for C-Eval we present is evaluated on development set.
}.

\textbf{Domain Evaluation.} 
To our knowledge, there is not any authoritative examinations on the domain of home renovation.
We construct a domain evaluation called EvalHome which covers three difficulty levels: domain fundamentals, domain expertise, and innovative design, from low to high difficulty, respectively.
Since multi-choice questions are a simple but good proxy to evaluate the potential of advanced abilities of our domain models, we construct all questions in a multi-choice format, resulting in 113 in total.
Table \ref{stat_evalhome} show the statistics of EvalHome.

\begin{table}[t!]
\caption{The statistics of EvalHome}
\begin{center}
\begin{tabular}{m{2cm}| p{1.5cm} p{1.5cm}} 
\hline 
\textbf{Category} & \textbf{\#Subclass} & \textbf{\#Questions} \\
\hline   
Domain fundamentals & 6  & 22  \\
\hline
Domain expertise & 17 & 87 \\
\hline
Innovative design  & 2  & 4 \\
\hline
\hline
TOTAL & 25  & 113  \\
\hline
\end{tabular}
\end{center}
\label{stat_evalhome}
\end{table}

\begin{table*}[t!]
\caption{Evaluation results on C-Eval and CMMLU after DAPT, and Pretrain Data-ratio represents the proportion of domain pre-training data and pre-training data. A ratio of '-' means that the model parameters are not updated. General pre-training data is randomly sampled from the general corpus we built \ref{Pre-trainingCorpus}.}
\small
\begin{center}
\begin{tabular}{lcccc} 
\hline 
\textbf{Base Model} & \textbf{PreTrain Data-ratio} & \textbf{C-Eval} & \textbf{CMMLU}\\
\hline
Baichuan-13B-Base & -- & 56.99 & 55.83 \\
\hline
\multirow{5}{*}{Baichuan-13B-Base} &1:0 &48.17 & 48.57 \\
 &1:1 &51.26 & 51.69\\
 &1:2 &46.12 & 47.75\\
 &1:5 &\textbf{54.42} & \textbf{52.75}\\
 &1:10 &49.26 & 48.59\\
\hline
\end{tabular}
\end{center}
\label{daptresult}
\end{table*}

\begin{table*}[t!]
\caption[The evaluation results of the model on EvalHome, C-Eval, and CMMLU after instruction alignment.] {SFT Data Ratio represents the proportion of domain instruction data and general instruction data. 
A ratio of '-' means that the model parameters are not updated. 
General instruction data is randomly drawn from 
Alpaca\_gpt4\_data\_zh\cite{peng2023instruction} and Belle\cite{BELLE}.}
\small
\begin{center}
\begin{tabular}{lcccc} 
\hline 
\textbf{Base Model for SFT} & \textbf{SFT Data Ratio} & \textbf{EvalHome} & \textbf{C-Eval} & \textbf{CMMLU}\\
\hline
Baichuan-13B-Chat & -- & 30.97 & 47.37 & 50.68\\
\hline
\multirow{5}{*}{Baichuan-13B-Base}
&1:0 &47.79 &\textbf{46.02} &\textbf{43.88}\\
&1:1 &50.44 &38.88 &40.18\\
&1:2 &44.24 &36.84 &39.89\\
&1:5 &36.28 &34.43 &36.84\\
&1:10 &\textbf{53.98} &38.52 &37.15\\
\hline
\multirow{5}{*}{Baichuan-13B-Base-DAPT(1:0)}
&1:0 &47.79 &\textbf{43.37} &\textbf{43.84}\\
&1:1 &46.01 &41.14 &39.03\\
&1:2 &47.79 &40.81 &39.92\\
&1:5 &\textbf{59.29} &39.90 &35.00\\
&1:10 &50.44 &35.01&35.83\\
\hline
\multirow{5}{*}{Baichuan-13B-Base-DAPT(1:5)}
&1:0 & 46.01 &\textbf{44.15} &\textbf{44.44}\\
&1:1 &47.79 &42.07 &41.33\\
&1:2 &48.67 &42.08 &39.60\\
&1:5 &\textbf{55.75} &38.08 &35.46\\
&1:10 &48.67 &37.79 &37.49\\
\hline
\multirow{5}{*}{Baichuan-13B-Chat}
&1:0 &37.16 &37.13 &34.62\\
&1:1 &51.21 &\textbf{42.01} &37.87\\
&1:2 &44.24 &41.12 &\textbf{39.72}\\
&1:5 &\textbf{60.17} &38.88 &37.26\\
&1:10 &45.13 &35.56 &36.99\\
\hline
\textbf{Base Model for MIP} & \textbf{MIP Data Ratio} & \textbf{EvalHome} & \textbf{C-Eval} & \textbf{CMMLU}\\
\hline
Baichuan-13B-Base  & 1:0 & \textbf{69.03} & \textbf{49.07} & \textbf{49.12}\\
\hline
\end{tabular}
\end{center}
\label{mainresult}
\end{table*}

\subsection{Results and Analysis}

\textbf{Data ratio result analysis} \quad The experimental results for the DAPT model on the general evaluation set are displayed in Table \ref{daptresult}.
We present the average scores on CEval and CMMLU, and the scores of each category-specific are shown in Table \ref{detailedscores}. 

Despite the addition of more general data at a 1:10 ratio, the DAPT model demonstrates the least general capability loss under a 1:5 data ratio scheme, and the average scores compared to the base model on the CEval and CMMLU evaluation sets decrease by 2.57 and 3.08 points respectively. This model is hereby denoted as Baichuan-13B-Base-DAPT (1:5).

Table \ref{mainresult} presents the experimental results of the domain-adapted model on the EvalHome and general evaluation sets. Four experiment groups were conducted, where Baichuan-13B-Base-DAPT (1:0) signifies a DAPT stage data ratio of 1:0. We can see that except for the Baihuan-13B-Base experiment, the other three experiments, Baihuan-13B-Base-DAPT (1:0), Baihuan-13B-Base-DAPT (1:5), and Baihuan-13B-Chat, all produced the best results on EvalHome under a 1:5 data ratio scheme.

Combining the experimental results of these two tables, we can draw a preliminary conclusion that a data ratio of 1:5 yields the best performance on our current base model and home renovation domain data.

During the instruction tuning phase, we observed a notable phenomenon: the model's scores on general capability evaluation sets decrease as more general instruction data is added. This could be attributed to the focus of the evaluation benchmarks, C-Eval and CMMLU, which primarily measure the model's specific knowledge that our general instruction data may not cover.

\textbf{Domain adaptation result analysis} \quad
It is observed that instruction tuning using the models after DAPT achieves optimal results of 59.29 and 55.75 on EvalHome respectively from Table \ref{mainresult}. There is a slight improvement compared to the Baichuan-13B-Base model without undergoing DAPT, which the highest score is 53.98. However, when instruction tuning is performed using the model Baichuan-13B-Chat that has been trained on instruction data, a higher score of 60.17 is obtained on EvalHome. And the model with different data ratios has achieved significant improvement compared to Baichuan-13B-Chat without updated parameters. This indicates that in our current domain scenario, post-DAPT instruction tuning doesn't significantly surpass direct domain adaptation in an instruction-aligned model. We speculate that this phenomenon may be due to the base model already encompassing a substantial volume of decoration-related data during pre-training.


Further, inspired by some research works \cite{Ext5,glm130b,galactica} showcasing the advantages of integrating downstream supervised datasets during pretraining, we attempted to incorporate downstream instruction data during the DAPT phase. This strategy is called MIP (Multi-Task Instruction Pre Training) in \cite{glm130b}, and we also follow this naming in this article. Owing to the constraints in training resources and time, we refrained from performing a detailed analysis of data ratios. Consequently, in the MIP stage, our training data consisted of domain pre-training data and domain instruction data exclusively, with no addition of a general corpus. Nonetheless, an unexpected score of 69.03 was achieved on EvalHome, as presented in the last row of Table \ref{mainresult}). Even more surprising is that this model outperforms all other models by not only achieving the highest score on EvalHome but also scoring higher on the two general capability evaluation benchmarks.

The finding indicates that incorporating downstream instruction data during the DAPT phase is beneficial, given the conditions of the current domain dataset and base models. Our future plans involve conducting more in-depth experiments with data ratios in the MIP stage.


\section{Discussion and Conclusions}
In this study, we introduce ChatHome, a large language model specifically designed for home renovation tasks. We conducted extensive experiments and performance evaluations, assessing domain-specific capabilities and general performance using various base models and data ratios.

However, our model still exhibits hallucination defects, and enhancing the assessment of its professional capabilities remains an ongoing area for improvement.In the future, we plan to conduct further experimental explorations, including the implementation of knowledge base and reinforcement learning techniques, to enhance ChatHome's performance.

\bibliographystyle{acl}
\bibliography{acl2015}
\appendix
\section{Appendix}
Prompts used for generating one-turn and multi-turn dialogues by GPT-4, examples of EvalHome and  the detail socres of C-Eval and CMMLU are presented in this appendix.

\begin{figure}[h]
\centering
\includegraphics[scale=0.8]{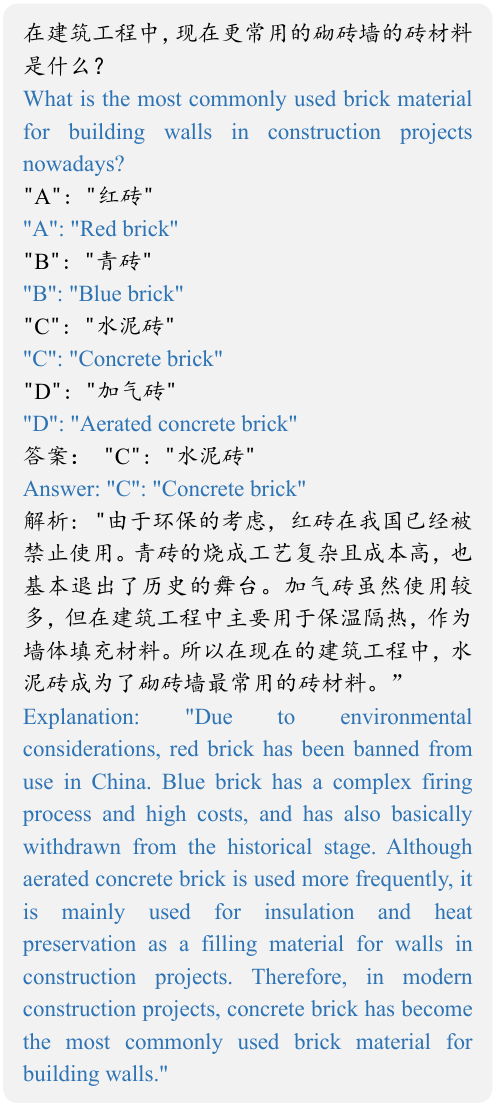} 
\caption{Example of EvalHome. English translations are shown below the corresponding Chinese for better readability.}
\label{example_evalhome}
\end{figure}

\begin{figure*}[b]
\centering
\includegraphics[scale=0.95]{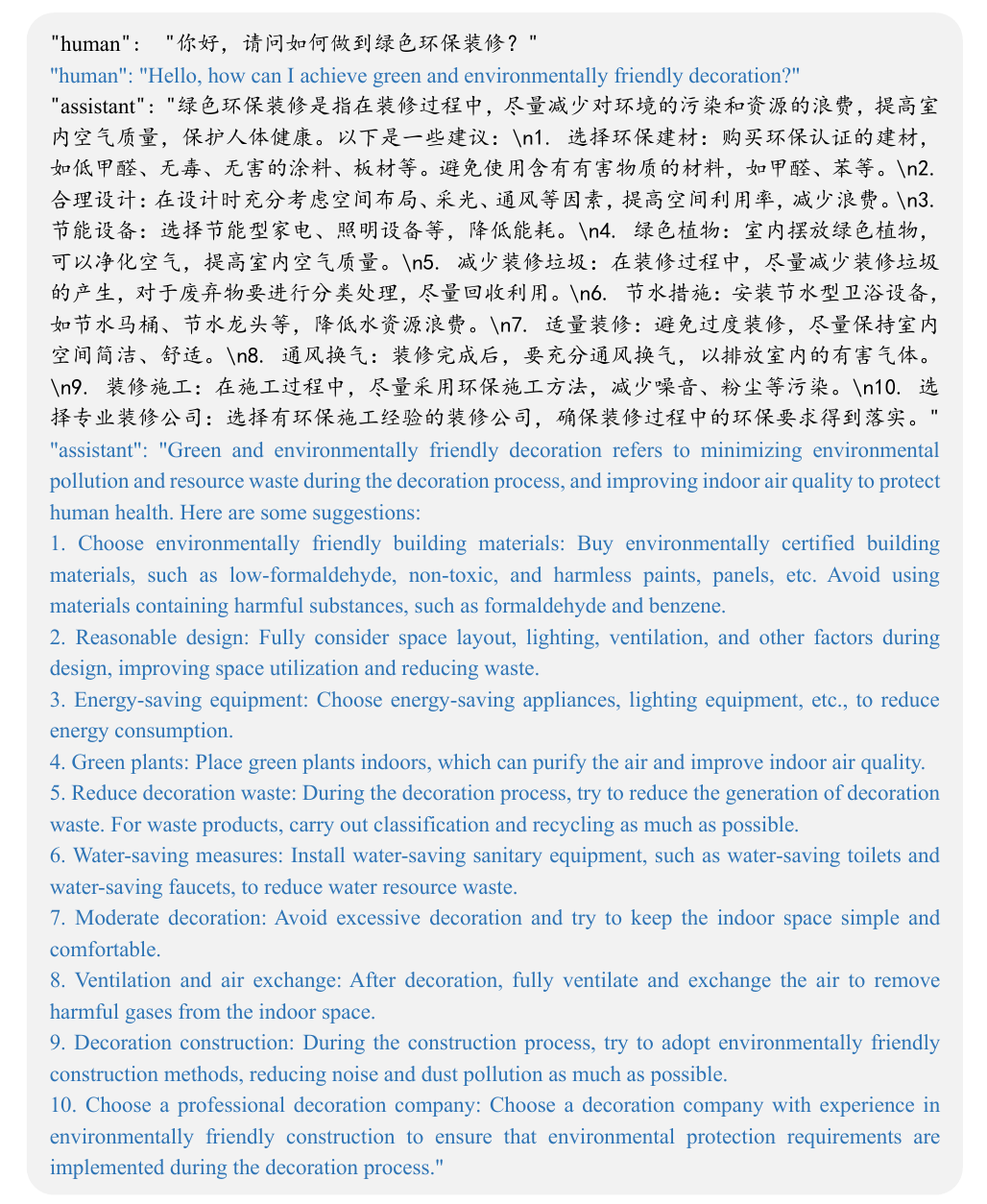} 
\caption{Example of one-turn dialogue. English translations are shown below the corresponding Chinese for better readability.}
\label{example_oneturn}
\end{figure*}

\begin{table*}[h]
\centering
\caption{Prompts used in the data generation process}
\begin{tabular}{m{1.5cm}|m{5cm}|m{8.5cm}}
\hline
\textbf{Type} & \textbf{Prompts} & \textbf{English translation}\\
\hline
one-turn dialogue & 

\begin{CJK*}{UTF8}{gkai}
你是一位资深家装领域从业者，现在请根据一段文本出5至20道题并给出对应的答案，出的题目要覆盖全部文本内容，出题的类别是以下40个类别之一：“行业标准、安装工程、工程验收、油漆工程、门窗工程......”。

具体文本是：(相关知识)

根据以上内容出题并给出答案，严格按照
\{['question':'xx','answer':'xx', 'category':'xx']\}
的格式提供问题和答案。
\end{CJK*}

& 
You are an experienced professional in the field of home decoration. Now, please create 5 to 20 questions and provide corresponding answers based on a given text. The questions should cover all the content of the text and be categorized into one of the following 40 categories: "Industry Standards, Installation Engineering, Engineering Acceptance, Painting Engineering, Door and Window Engineering..."

The specific text is: (Relevant Knowledge)

Please provide questions and answers according to the format [’question’:’xx’,’answer’:’xx’, ’category’:’xx’].\\
\hline

multi-turn dialogue & 
\begin{CJK*}{UTF8}{gkai}
请基于以下背景信息，生成一个user和你的对话，需要满足：1、用户不会称呼你的名字。2、你的每一次回答，都和你日常回答一样，都是高质量的非常详尽的回答。3、用户有可能会基于你的回答提出指令。4、对话内容需要尽可能的涵盖背景信息中的内容

背景信息: (相关知识) 

以下是生成的对话:
\end{CJK*}
& 
Please generate a dialogue between a user and you based on the following background information, meeting the following requirements: 1) The user will not address your name. 2) Your every answer should be of high quality and very detailed, the same as your usual answers. 3) The user may issue instructions based on your answers. 4) The dialogue content should cover as much of the background information as possible.

Background information: (Relevant Knowledge)

Here is the generated dialogue: \\
\hline
multiple-choice question
&
\begin{CJK*}{UTF8}{gkai}
你是一位资深家装领域从业者，请根据给定的知识设计一道单选题或判断题。单选题应包括A、B、C、D四个选项，其中只有一个选项是正确的；判断题只需判断对或错，以AB两个选项表示。最后请输出正确选项，并提供理由。

具体知识是：(相关知识)

结果严格按照\{'question':'单选题或者判断题题目','question\_type':'单选或判断','candidate\_options':'此项是候选项内容，也以json格式输出，其中key是选项名，value是选项内容', 'answer':\{'correct\_option':'输出唯一正确的选项名，应与理由保持一致','reason':'选项正确的理由'\}\}的json格式输出。
\end{CJK*} 
& 
You are an experienced professional in the field of home decoration. Please design a multiple-choice question or a judgment question based on the given knowledge. 
The multiple-choice question should include options A, B, C, and D, with only one option being correct. 
The judgment question only needs to judge whether it is right or wrong, represented by options A and B. 
Finally, please output the correct option and provide the reason.

Specific knowledge is: (Relevant Knowledge)

The output should strictly follow the JSON format \{  
 'question': 'Multiple-choice or judgment question title',  
 'question\_type': 'multiple-choice or judgment',  
 'candidate\_options': 'This item is the content of the candidates, also output in JSON format, where the key is the option name and the value is the option content',  
 'answer': \{'correct\_option': 'Output the name of the only correct option, which should be consistent with the reason',  
   'reason': 'The reason why the option is correct'  
 \}  
\}\\
\hline

\end{tabular}
\label{prompts}
\end{table*}




\begin{figure*}[ht]
\centering
\setlength{\abovecaptionskip}{0.1cm} 
\includegraphics[scale=0.95]{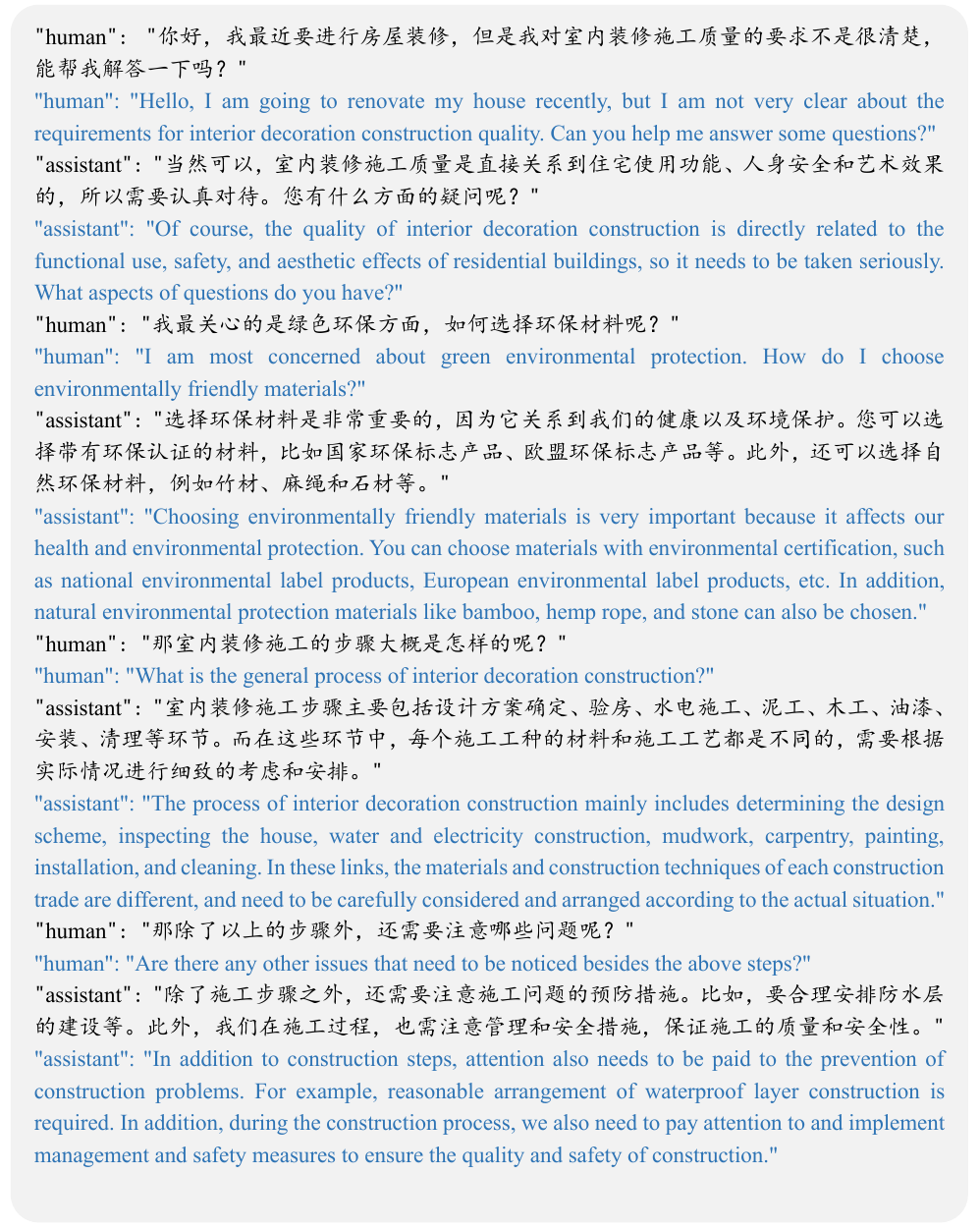} 
\caption{Example of multi-turn dialogue. English translations are shown below the corresponding Chinese for better readability.}
\label{example_multiturn}
\end{figure*}

\begin{sidewaystable*}[b!]
\caption{Detailed evaluation score on C-Eval and CMMLU}
\small
\begin{center}
\resizebox{1.0\textwidth}{!}{
\begin{tabular}{llccccccccccc} 
\hline 
\multirow{2}{*}{\textbf{Base Model for DAPT}} & \multirow{2}{*}{\textbf{PreTrain Data-ratio}} & \multicolumn{5}{c}{\textbf{C-Eval}} & \multicolumn{6}{c}{\textbf{CMMLU}}
\\
 &  & \textbf{Avg} & \textbf{STEM} & \textbf{Social Science} & \textbf{Humanities} & \textbf{Others} & \textbf{Avg} & \textbf{STEM} & \textbf{Social Science} & \textbf{Humanities} & \textbf{Others} &\textbf{China specific} \\
\hline
Baichuan-13B-Base & -- & 56.99 & 44.21 & 71.77 & 60.65 & 51.92 & 55.83 &42.61 & 60.47 & 61.47 & 59.13 &56.81\\
\hline
\multirow{5}{*}{Baichuan-13B-Base} &1:0 &48.17 &37.95 &56.20 &52.82 &45.72 &48.57 &37.60 &52.76 &52.92 &51.07 &48.68 \\
 &1:1 &51.26 & 39.89 & 64.13 & 54.21 & 46.79 & 51.69 & 39.09 & 56.08 & 56.47 & 55.38 &51.10\\
 &1:2 &46.12 & 33.76 & 58.28 & 50.34 & 42.08 & 47.75 & 35.75 & 53.42 & 51.29 & 49.97 &48.08\\
 &1:5 &54.42 & 38.87 & 69.55 & 59.50 & 49.76 & 52.75 & 40.40 & 58.03 & 58.14 & 54.34 &53.90\\
 &1:10 &49.26 & 40.54 & 63.65 & 50.60 & 42.23 & 48.59 & 37.86 & 53.78 & 50.12 & 51.81 &49.65\\
\hline
\multirow{1}{*}{\textbf{Base Model for SFT}} & \multirow{1}{*}{\textbf{SFT Data-ratio}} & \multicolumn{5}{c}{-} & \multicolumn{6}{c}{\textbf{-}} \\
\hline
\multirow{5}{*}{Baichuan-13B-Base}
&1:0 &46.02 &35.38 &54.88 &49.43 &44.38 &43.88 &32.75 &48.25 &49.39 &45.28 &44.37\\
&1:1 &38.88 &35.38 &41.04 &41.93 &38.14 &40.18&34.22 &42.03 &42.31 &42.37 &40.29\\
&1:2 &36.84 &33.05 &40.25 &41.55 &32.52 &39.89 &31.59 &42.93 &42.15 &42.90 &41.65\\
&1:5 &34.43 &30.65 &37.70 &35.84 &33.53 &36.84 &30.26 &39.09 &38.94 &39.16 &36.74\\
&1:10 &38.52 &35.02 &42.98 &39.72 &36.35 &37.15 &30.75 &39.09 &40.10 &39.04 &37.68\\
\hline
\multirow{5}{*}{Baichuan-13B-Base-DAPT(1:0)}
&1:0 &43.37 &35.35 &47.74 &48.69 &41.68 &43.84 &33.35 &48.68 &48.21 &44.83 &45.26\\
&1:1 &41.14 &35.79 &43.70 &47.78 &37.27 &39.03 &32.11 &41.65 &41.99 &40.45 &41.07\\
&1:2 &40.81 &34.88 &43.78 &46.00 &38.60 &39.92 &32.49 &42.85 &43.32 &41.09 &40.57\\
&1:5 &39.90 &35.81 &41.16 &45.03 &37.60 &35.00 &30.34 &35.65 &37.45 &37.22 &34.02\\
&1:10 &35.01 &33.03 &38.62 &35.78 &32.62 &35.83 &30.13 &38.26 &36.04 &38.57 &36.08\\
\hline
\multirow{5}{*}{Baichuan-13B-Base-DAPT(1:5)}
&1:0 &44.15 &36.10 &50.17 &49.62 &40.70 &44.44 &35.08 &47.89 &49.48 &45.63 &45.67\\
&1:1 &42.07 &37.82 &41.86 &46.01 &42.59 &41.33 &33.33 &44.33 &44.91 &42.89 &42.80\\
&1:2 &42.08 &36.43 &44.32 &48.76 &38.83 &39.60 &31.63 &42.64 &43.49 &40.81 &40.22\\
&1:5 &38.08 &33.21 &40.34 &41.91 &36.84 &35.46 &29.88 &37.35 &36.35 &38.25 &36.43\\
&1:10 &37.79 &34.84 &38.49 &41.26 &36.57 &37.49 &29.55 &40.26 &40.31 &39.98 &35.62\\
\hline
\multirow{5}{*}{Baichuan-13B-Chat}
&1:0 &37.13 &32.55 &36.95 &42.33 &36.68 &34.62 &28.67 &35.96 &37.26 &37.11 &35.96\\
&1:1 &42.01 &34.46 &47.48 &47.41 &38.68 &37.87 &31.75 &39.00 &40.88 &40.55 &37.85\\
&1:2 &41.12 &34.14 &46.73 &45.75 &37.84 &39.72 &33.42 &41.71 &42.05 &41.94 &40.17\\
&1:5 &38.88 &34.64 &40.58 &37.87 &42.43 &37.26 &31.40 &37.84 &40.25 &40.44 &36.82\\
&1:10 &35.56 &32.87 &37.81 &33.08 &38.47 &36.99 &29.98 &40.04 &38.66 &39.03 &36.99\\
\hline
Baichuan-13B-Chat & -- & 47.37 & 43.28 & 54.47 & 49.27 & 42.44 & 50.68 & 39.1 & 55.29 & 54.82 & 53.45 &52.26\\
\hline
\textbf{Base Model for MIP} & \textbf{MIP Data Ratio} & \multicolumn{5}{c}{-} & \multicolumn{6}{c}{\textbf{-}}\\
\hline
Baichuan-13B-Base  & 1:0 & 49.07 & 35.76 & 64.88 & 51.39 & 44.24 & 49.12 & 36.82 & 54.57 & 53.52 & 51.24 &49.09 \\
\hline
\end{tabular}
}
\end{center}
\label{detailedscores}
\end{sidewaystable*}

\end{document}